\documentclass[10pt,twocolumn,letterpaper]{article}
\usepackage{cvpr}

\usepackage[scaled=.92]{helvet}
\usepackage{times}
\usepackage{graphicx}
\usepackage{parskip}
\usepackage[labelfont=bf,textfont=it]{caption}
\usepackage{microtype}
\usepackage{amssymb}
\usepackage{amsmath}
\usepackage{mathptmx}
\usepackage{mathrsfs}
\usepackage{fixltx2e}
\usepackage{wrapfig}
\usepackage{xspace}
\usepackage{float}
\usepackage{filecontents}
\DeclareMathAlphabet{\mathcal}{T1}{pzc}{mb}{it}
\RequirePackage[T1]{fontenc}
\RequirePackage[override]{cmtt}

\usepackage{units}
\usepackage{tabu}
\usepackage{booktabs}
\usepackage{color}
\usepackage{soul}
\usepackage{subcaption}

\def\figurePath{}
\def\myfigure#1#2{\begin{figure}[htpb]\centering\includegraphics*[width = \linewidth]{\figurePath#1}\caption{#2 }\label{fig:#1}\end{figure}}
\def\mycfigure#1#2{\begin{figure*}[t]\centering\includegraphics*[clip, width = \linewidth]{\figurePath#1}\caption{#2 }\label{fig:#1}\vspace{-0.5\baselineskip}\end{figure*}}

\newcommand{\NN}{{\sc NN}\xspace}
\newcommand{\LINEAR}{{\sc Linear}\xspace}
\newcommand{\OPT}{{\sc Opt}\xspace}
\newcommand{\DQ}{{\sc DQ}\xspace}
\newcommand{\GAUSS}{{\sc GAUSS}}
\newcommand{\HEVC}{{\sc HEVC}\xspace}

\newcommand{\PSNR}{{\sc PSNR}\xspace}
\newcommand{\SSIMSS}{{\sc SSIM\textsubscript{2D}}\xspace}
\newcommand{\SSIMSSA}{{\sc SSIM\textsubscript{2D$\times$1D}}\xspace}
\newcommand{\SSIMSAA}{{\sc SSIM\textsubscript{3D}}\xspace}
\newcommand{\VDP}{{\sc HDR-VDP-2}\xspace}
\newcommand{\VQM}{{\sc VQM}\xspace}

\newcommand{\MSSIM}{{\sc MS-SSIM}\xspace}
\newcommand{\GMSD}{{\sc GMSD}\xspace}
\newcommand{\CHEN}{{\sc SIQM}\xspace}
\newcommand{\BATTISTI}{{\sc 3DSwIM}\xspace}
\newcommand{\SANDIC}{{\sc MP-PSNR}\xspace}
\newcommand{\SILVA}{{\sc StSD\textsubscript{LC}}\xspace}

\newcommand{\JODs}{{\sc JODs}\xspace}
\newcommand{\JOD}{{\sc JOD}\xspace}

\def\mysection#1#2{\section{#1}\label{sec:#2}}
\def\mysubsection#1#2{\subsection{#1}\label{sec:#2}}

\definecolor{piotrcolor}{rgb}{0.1,0.6,0.2}

\definecolor{rafalcolor}{rgb}{0.6,0.2,0.2}

\definecolor{Deniscolor}{rgb}{0.6,0.2,0.6}

\definecolor{marekcolor}{rgb}{0.1,0.25,0.65}

\definecolor{karolcolor}{rgb}{0.7,0.25,0.3}

\definecolor{vamsicolor}{rgb}{1.0,0.65,0}
\newcommand{\vamsi}[1]{\textcolor{vamsicolor}{Vamsi: #1}}

\definecolor{changedcolor}{rgb}{1.0,0.0,0.0}
\newcommand{\changed}[1]{\textcolor{changedcolor}{#1}}

\renewcommand{\st}[1]{}
\renewcommand{\vamsi}[1]{}
\renewcommand{\changed}[1]{#1}

\DeclareGraphicsExtensions{.png,.pdf,.ai,.psd}
\usepackage[pagebackref=true,breaklinks=true,letterpaper=true,colorlinks,bookmarks=false]{hyperref}

\cvprfinalcopy 

\def\cvprPaperID{28} 

\ifcvprfinal\fi
\begin{document}

\title{Towards a quality metric for dense light fields}
\author{Vamsi Kiran Adhikarla$^{1}$
\and
Marek Vinkler$^{1}$
\and
Denis Sumin$^{1}$
\and
Rafa\l{} K. Mantiuk$^{3}$
\and
Karol Myszkowski$^{1}$
\and
Hans-Peter Seidel$^{1}$
\and
Piotr Didyk$^{1,2}$
}



\twocolumn
[
	\begin{@twocolumnfalse}
	\maketitle
	\vspace{-1.0\baselineskip}
	\begin{center}	
	$^{1}$MPI Informatik \quad $^{2}$Saarland University, MMCI \quad $^{3}$The Computer Laboratory, University of Cambridge
	\end{center}
	\vspace{2\baselineskip}
	\end{@twocolumnfalse}
]
\thispagestyle{empty}

\begin{abstract}

Light fields become a popular representation of three-dimensional
scenes, and there is interest in their processing, resampling, and
compression. As those operations often result in loss of quality, there
is a need to quantify it. In this work, we collect a new dataset of dense reference and
distorted light fields as well as the corresponding quality scores which are scaled in
perceptual units. The scores were acquired in a subjective experiment using
an interactive light-field viewing setup. The dataset contains typical
artifacts that occur in light-field processing chain due to light-field reconstruction, multi-view compression, and limitations of automultiscopic
displays. We test a number of existing objective quality
metrics to determine how well they can predict the quality of light fields. We find that the existing image quality metrics provide good
measures of light-field quality, but require dense reference light-
fields for optimal performance. For more complex tasks of comparing two distorted light fields, their performance drops significantly, which reveals the need for new, light-field-specific metrics.


%
%
\end{abstract}


\mysection{Introduction}{Introduction}

A light field can be seen as a generalization of a 2D image, which
encodes most of the depth cues and allows to render a
scene simulating arbitrary optics (e.g., defocus
blur) \cite{Levoy1996}. It is a convenient representation for
multiscotopic and \changed{light-field displays}~\cite{Wetzstein2012},
but also attractive format for capturing high-quality cinematographic
content, offering new editing possibilities in post-production \cite{LytroCinema}. Due the enormous storage requirements, light fields are usually
sparsely sampled in spatial and angular dimensions, stored using lossy
compression, and reconstructed later. \changed{It is unclear how the distortions introduced on the way affect the perceived quality.}


Similar problems have been addressed for 2D images, videos, and sparse multiview content. Many quality metrics have been designed to predict perceived differences between various versions of the \changed{same content}~\cite{Aydin2010}. However, measuring quality for dense light fields still remains a complex task. While several works applied the existing metrics to \changed{such content} \cite{Higa2013,Dansereau2013}, their performance has never been systematically evaluated in this context. One of the challenges is acquiring dense light-field data to validate a metric. Wide baselines as in multi-camera rigs~\cite{Wilburn2005} need to be considered, and the reference light fields should be sufficiently dense to avoid uncontrolled visual artifacts. Obtaining human responses for light-field distortions is also difficult due to current display limitations. This work is an attempt to overcome these problems by first building a new dense light-field dataset which is suitable for testing quality metrics, and second, using a custom light-field viewing setup to obtain the quality judgments for this dataset. The collected subjective scores are used to evaluate the performance of existing metrics in the context of dense light fields. 

We focus on light-field-specific angular effects akin to motion parallax, complex surface appearance, and binocular vision that arise in free viewing experience. To capture a rich variability over these effects and make quality scaling in our perceptual experiments tractable, we design fourteen real and synthetic scenes and introduce light-field distortions that are specific to light-field reconstruction, compression, and display. We then run a pair-wise comparison experiment over light-field pairs, and derive perceptual scaling of differences between original and distorted stimuli. This allows us to investigate the suitability of a broad spectrum of existing image, video, and multiview quality metrics to predict such perceptual scaling. We also propose simple extensions of selected metrics to capture the angular aspects of light-field perception. While the original metrics are not meant for light fields, our results show that they can be used in this context, given a dense light field as the reference. We also demonstrate that the robustness of such metrics predictions drops when evaluating the quality between two distorted light fields.
The main contributions of this work are:
\begin{itemize}
\vspace{-0.5em}
\item a publicly available dense light-fields dataset that is designed for training and evaluating quality metrics;
\vspace{-0.5em}
\item a perceptual experiment that provides human quality judgments for several typical light-field distortions;
\vspace{-0.5em}
\item an evaluation, analysis, and extensions of existing quality metrics in the context of light fields;
\vspace{-0.5em}
\item identified challenges of quality assessment in light fields, such as the need for a high quality reference. 
\end{itemize}

\mysection{Previous works}{PreviousWork}
In this section, we provide an overview of existing datasets for light fields as well as the experiments that measure the perceived distortions in various types of content. 

\textbf{Light-field datasets: }
There are several publicly available light-field datasets. The most popular ones are: 4D light-field dataset ~\cite{Wanner13} containing seven synthetic scenes and five real-world scenes, Stanford archive \cite{Stan2008} with twenty 4D light fields, and Disney 3D light-field dataset \cite{Kim2013} containing five scenes. Although the first two datasets provide a good quality and reasonable number of light fields, they are captured over very narrow baselines that are insufficient for the new generation autostereoscopic displays. The Disney dataset provides high spatio-angular resolution light fields; however, they are few and do not have consistent spatial and angular resolution, which makes it difficult to use in quality evaluations. In the context of quality evaluation of 3D light fields, three real-world light fields are provided in the IRCCyN/IVC DIBR Images database \cite{Bosc2011}. These contain several scenes captured along a wide baseline at the cost of reduced angular resolution. Tamboli~et~al. \cite{Tamboli16} provided 360$^{\circ}$ round table shots of three scenes that are used for quality evaluation on a 3D light-field display. These are rather simple scenes with single objects and the images contain a lot of noise. In our work, we provide first consistent dataset of dense, complex-scene light fields with large appearance variation. We use the dataset for training and evaluating quality metrics. The database cane also serve as a ground truth for automultiscopic displays.

\textbf{Metrics and experiments: }
Because of their proven efficiency on 2D images, 2D objective metrics are viable candidates for evaluating light-field quality. Yasakethu et al. \cite{Yasakethu08} tested the suitability of objective measures -- Structural SIMilarity (SSIM)~\cite{Wang2004}, Peak Signal-to-Noise Ratio (PSNR) and Video Quality Metric (VQM)~\cite{Pinson2004} for quality assessment for stereoscopic and 2D+Depth videos that are compressed at different bitrates. They carried out subjective experiments on an autostereoscopic display and showed that 2D metrics can be used separately on each view to assess 3D video quality. They used few sequences and studied only compression artifacts. \changed{Several metrics have been proposed to determine the quality of synthesized views from multiview images. Bosc et al.{~\cite{Bosc2011}} advocated two measures for assessing the quality of synthesized views. However, they did not conduct thorough subjective studies. Solh et al.{~\cite{Solh2009}} presented a metric for quantifying the geometric and photometric distortions in multiview acquisition. Bosc et al.{~\cite{Bosc2013}} suggested a method to asses the quality of virtual synthesized views in the context of multiview video. Battisti et al.{~\cite{Battisti2015}} proposed more sophisticated framework for evaluating the quality of depth image based rendering techniques by comparing the statistical features of wavelet subbands and used image registration and skin detection steps for additional optimization. Sandic et al.{~\cite{Sandic2016}} exploited multi-scale pyramid decompositions with morphological filters for obtaining the quality of intermediate views and showed that they achieve significantly higher correlation with subjective scores. These methods form a class of metrics specific to view-interpolation artifacts, and 2D stimuli containing the interpolated views are used for subjective experiments}. 

Vangorp et al. \cite{Vangorp2011} ran a psychophysical study to account for the plausibility of visual artifacts associated with view interpolation methods. They considered such artifacts as a function of different number of input images; however, they limited their study to monocular viewing and Lambertian surfaces. An experiment was also performed for precomputed videos, so that the impact of user's interaction and dynamic aspects of free viewing could be judged. More recently, this work was extended to transitions between videos \cite{Tompkin2013}. Similar studies were also performed in the context of panoramas \cite{Morvan2009}. Tamboli et al.~\cite{Tamboli16} conducted subjective studies on a 3D light-field display. Users were asked to judge the quality as perceived from different viewing locations in front of the display and the scores were averaged over all locations. The user could rate the quality only from a certain viewing position. Moreover, they only considered three distinct scenes. We believe that, for inferring a light-field quality, all the views should be taken into account at the same time.

\textbf{Light-field displays: }
\changed{Our work focuses on wide-baseline 3D light fields which enable perfect simulation of stereoscopic viewing and continuous horizontal motion parallax crucial for new light-field displays. Although many light-field display designs exist{~\cite{Masia2013}}, including more advanced ones that provide focus cues{~\cite{Maimone2013}}, they suffer from several drawbacks such as limited field of view, discontinous motion parallax, visible crosstalk, and limited depth budget. Several strategies have been proposed to minimize these artifacts by filtering the content~\cite{Zwicker2006,Du2014} and manipulating depth~\cite{Lang2010,Didyk2012,Masia2013}. However, display designs that enable displaying reference light fields for quality measurements are still unavailable.}

\mycfigure{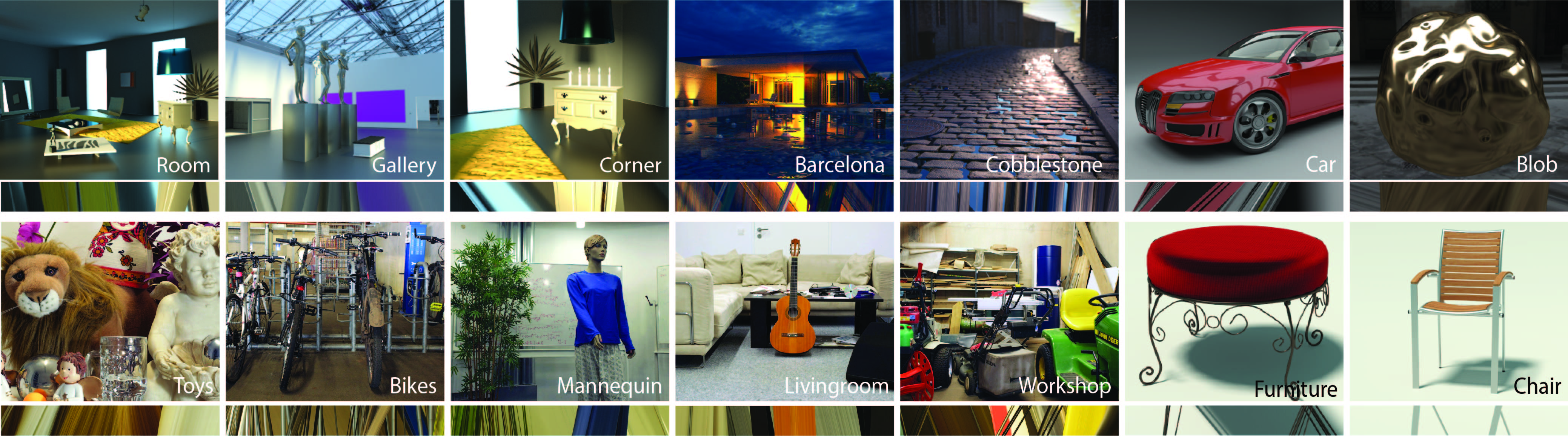}{
Representative images of all light fields in our collection. Below each image representative EPIs are presented.
}
\mysection{Data collection}{DataCollection}
Our dataset consist of light fields which are parameterized using two parallel planes~\cite{Levoy1996}. We consider only horizontal motion parallax that can be described using one plane and a line that is parallel to it. 

\setlength{\intextsep}{5pt}
\setlength{\columnsep}{10pt}
\begin{wrapfigure}{r}{0.5\linewidth}
    \includegraphics[width=\linewidth]{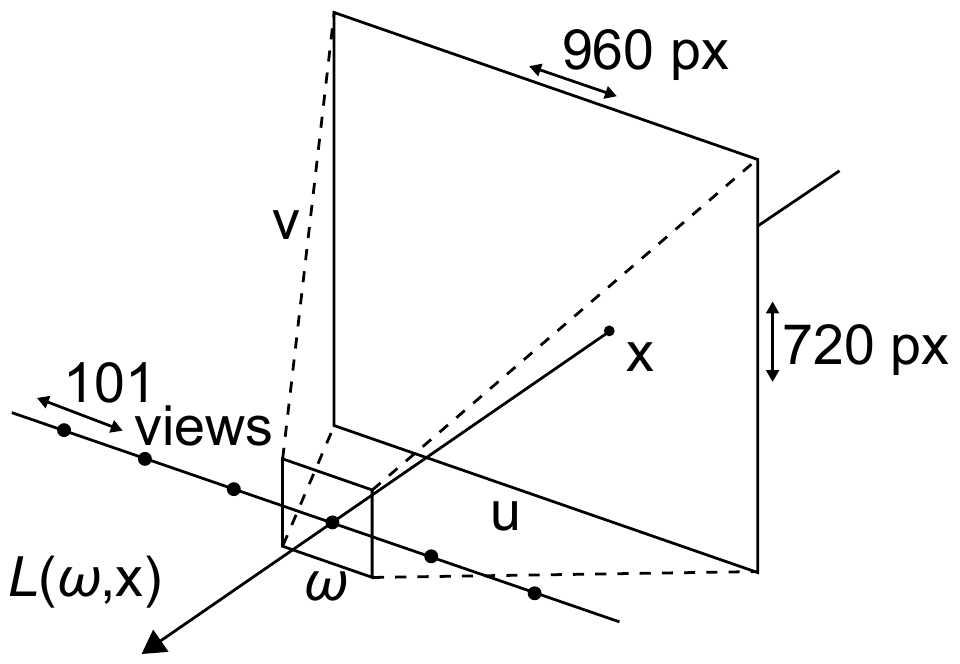}
\end{wrapfigure}
More formally, we denote our light fields as $L(\mathbf \omega, \mathbf x)\in(\mathbb R \times \mathbb R^2)\rightarrow\mathbb R^3$, where $\mathbf \omega$ is a position on the line, and $\mathbf x$ is a position on the plane. We refer to them as \emph{angular} and \emph{spatial} coordinates, respectively. In practice, $\mathbf \omega$ describes a position of the viewer, and $\mathbf x$ is a coordinate of the observed image. Below, we describe the acquisition of our light fields.

\mysubsection{Scenes}{Scenessec}
We designed and rendered nine synthetic and captured five real-world scenes (Figure~\ref{fig:Scenes.jpg}). They span a large variety of different conditions, \eg outdoor/indoor, daylight/night etc. They also contain objects with large range of different appearance properties. The scene objects distribution in depth is widely varied to study the artifacts resulting from disocclusions and depth discontinuities. For capturing real-world scenes, we used a one-meter long motorized linear stage with Canon EOS 5D Mark II camera and 50\,mm and 28\,mm lenses. \changed{After capturing all views, we performed lens distortion correction using \changed{PTLens~\cite{PTlens}}, estimated the camera poses using \changed{Voodoo camera tracker~\cite{Voodoo}}, and rectified all the images \changed{using the baseline drawn from the first to last camera using the approach in~\cite{Fusiello2000}}.}
For rendered images, we used cameras with off-axis asymmetric frustums. For real-world scenes, the same effect was achieved by applying horizontal shift to the individual views.
All the light fields are of identical spatial and angular resolution (960$\times$720$\times$101). The angular resolution was chosen high enough to avoid visible angular aliasing. This was achieved by assuring that the maximum on-screen disparities between consecutive views are around 1\,pixel. To guarantee a comfortable viewing, the total disparity range during the presentation was limited to 0.2 visual degree~\cite{Shibata2011}.

\mysubsection{Distortions}{Distortions}
We considered typical light-field distortions that are specific to transmission, reconstruction, and display. For each distortion, we generated multiple light fields by varying the distortion severity level. \changed{The exact levels were chosen to keep the differences between two consecutive levels small and similar. To this end, we conducted a small pilot study with 10 distortion levels, and then, selected the final levels manually.}

\textbf{Transmission:} To transmit the light-field data, an efficient data compression algorithm is highly required. We consider well-known 3D extension of HEVC encoder \cite{GTech2016}. The light-field views are encoded into a bit stream at various quantization steps, and then, decoded back from the bit stream using the 3D-HEVC coder. We chose the following quantization steps: \{25, 29, 33, 37, 41, 45\}.

\textbf{Reconstruction:} Light-field reconstruction techniques are used to recover a dense light field from sparse view samples. They interpolate the missing views using several techniques which alter the nature and appearance of the distortion. We chose the distortions resulting from linear (\LINEAR) and nearest neighbor (\NN) interpolation, as well as image warping using optical flow estimation (\OPT). We also investigated the impact of using quantized depth maps (\DQ). All the distortions are parametrized by the angular subsampling factor $k$ (the distortion severity) that defines the angular resolution of the light field prior to applying the reconstruction technique. We considered $k \in \{2,5,8,11,18,25\}$. The linear filter reconstructs dense light field by blending the reference views, and the \NN method clones the closest reference view. For \OPT method, we used the TV-L1 optical flow~\cite{Sanchez2013}, and apply an image-warping technique~\cite{Brox2004} to synthesize in-between views. For \DQ, we considered ground-truth depth map which is quantized using 8 discrete levels. Then, we used the same image-warping technique~\cite{Brox2004} to reconstruct the light field. As this distortion requires ground-truth depth information, it is only applied to the synthetic scenes.

\textbf{Display:} As an example of multiview autostereoscopic display artifacts, we chose a crosstalk between adjacent views, which can be modeled using a Gaussian blur in angular domain \cite{Liu2015}. Consequently, we include such artifacts into our dataset (\GAUSS). In particular, we considered the same angular subsampling parameters used in light-field reconstruction distortions and created hypothetical displays with corresponding number of views. The upsampling to higher resolution light field was achieved by using the display crosstalk model.

Four different distortions with all severity levels were applied to every scene. To all synthetic scenes we apply \NN, \LINEAR, \OPT, and \DQ. For all real-world scenes, we used \NN, \OPT, \GAUSS, and \HEVC. Including original light fields, our database consists of 350 different light fields and it is available online~\cite{MPILF}. The examples of resulting artifacts are presented in Figure~\ref{fig:Distortions}. Please refer to supplemental materials for the whole light-field dataset.

\myfigure{Distortions}{
Examples of distortions introduced to our light field for one of our scenes ({\sc Barcelona}). The images visualize central EPI of each of the distorted light fields and the enlarged portion of it is shown in the bottom row.
}


\mysection{Experiment}{Experiment}
To acquire subjective quality scores that enable  both training and testing different quality metrics, we performed a large-scale subjective experiment.

\textbf{Equipment:}
\changed{To simulate stereoscopic viewing with high-quality motion parallax, we used on our own setup} ({Figure~\ref{fig:ExperimentSetup}) that consists of ASUS VG278 $27\,''$ Full HD 120\,Hz LCD desktop monitor and NVIDIA 3D Vision 2 Kit for displaying stereoscopic images. Motion parallax was reproduced using a custom head tracking in which a small LED headlamp was tracked using a Logitech HD C920 Pro webcam (refer to the supplemental video). 
The head tracking allowed the participants to view light fields in an unconstrained manner. The viewing distance was approximately 60\,cm, and users could move their heads along a baseline of 20\,cm in the direction parallel to the screen plane. 
The eye accommodation was fixed to the screen and did not change with eye vergence. \changed{The display was operated at the full brightness to minimize the effect of luminance on depth perception{~\cite{Didyk2012}}.}

\myfigure{ExperimentSetup}{Experiment session: viewer's position is tracked using a head lamp and a webcam, a pair of NVIDIA 3D Vision 2 Kit active glasses provides stereoscopic viewing.}

\textbf{Stimuli:}
Each stimuli was a pair of light fields. As our scaling procedure used for obtaining quality scores (Section~\ref{sec:Analysis}) can handle an incomplete set of comparisons and prefers when more comparisons are made for pairs of similar quality \cite{Silverstein2001}, each pair consisted of light fields with neighboring severity levels of the same distortion type. This results in 336 different stimuli which were presented stereoscopically.

\textbf{Task:}
We experimented with direct rating methods, such as ACR \cite{ITU-T-P.9102008}, in order to measure Mean-Opinion-Scores of the distorted images. However, we found these methods to be insensitive to subtle but noticeable degradation of quality. Also participants found the direct rating task difficult. Therefore, we decided to use a more sensitive pair-wise comparison method with a two-alternative-forced-choice. In each trial, the participants were shown a pair of light fields side-by-side, and the task was to indicate the light field that a user ``would prefer to see on a 3-dimensional display''. 
Participants were given unlimited time to investigate the light fields, but they were allowed to give their response only after 80\% of all perspective images were seen. The order of the light-fields pairs as well as their placement on the screen were randomized. Before each session, the participants were provided with a form summarizing the task, and a training session was conducted to familiarize participants with the experiment.

\textbf{Participants:}
Forty participants took part in the test, including both male (20) and female (20) aged 24--40 with normal or corrected-to-normal vision. Each subject performed the test in three sessions within one week. In one session, the participants saw 120--180 light-field pairs consisting of all the test conditions, but for a subset of the scenes. 
For a given subject, two test sessions were allowed during a single day, and these were separated by at least an hour of break.

\mysection{Analysis of subjective data}{Analysis}

The results of pair-wise comparison experiment are usually scaled in just-noticeable-differences (JNDs). We observed that considering measured differences as ``noticeable'' leads to incorrect interpretation of the experimental results. Two stimuli are 1\,JND apart if 75\% of observers can see the difference between them. However, our experimental question was not whether observers can tell if the light fields are different, but rather which one has higher quality. As shown in Figure~\ref{fig:jod_vs_jnd}, a pair of stimuli could be noticeably different from each other (JND>1), but they could appear to have the same quality. For that reason, we denote measured values as just-objectionable-differences (\JODs). These units quantify the quality difference in relation to the perfect reference image. Note that the measure of \JOD is more similar to visual equivalence \cite{Ramanarayanan2007a} or to the quality expressed as a difference-mean-opinion-score (DMOS) rather than to JNDs.

\myfigure{jod_vs_jnd}{Illustration of the difference between just-objectionable-differences (JODs) and just-noticeable-differences (JNDs). The image affected by blur and noise may appear to be similarly degraded in comparison to the reference image (the same JOD), but they are noticeably different and therefore several JNDs apart. The mapping between JODs and JNDs can be very complex and the relation shown in this plot using Cartesian and polar coordinates is just for illustration purposes. 
}

To scale the results in \JOD units we used a Bayesian method based on the method of Silverstein and Farrell~\cite{Silverstein2001}. It employs a maximum-likelihood-estimator to maximize the probability that the collected data explains \JOD-scaled quality scores under the Thurstone Case V assumptions~\cite{RAFALC}. The optimization procedure finds a quality value for each pair of light fields that maximizes the likelihood modeled by the binomial distribution. Unlike standard scaling procedures, the Bayesian approach robustly scales pairs of conditions for which there is unanimous agreement. Such pairs are common when a large number of conditions are compared. It can also scale the result of an incomplete and imbalanced pair-wise design, when not all the pairs are compared and some are compared more often. As the pair-wise comparisons provide relative quality information, the \JOD values are relative. To maintain consistency across the scenes, we fix the starting point of the \JOD scale at 0 for different distortions and thus the quality degradation results in negative \JOD values.

The results of the subjective quality assessment experiment are shown in Figure~\ref{fig:exp_results}. The error bars represent 95\% confidence intervals, relative to the reference light field, computed by bootstrapping by sampling with replacement.
\mycfigure{exp_results}{
The results of the subjective quality assessment experiment. The distortion level indicates the distortion severity: 0--reference 6--severest distortion level. JOD is the scaled subjective quality value. The error bars denote 95\% confidence interval. The bars are horizontally displaced to avoid overlapping. The scene names are indicated in the corner of each plot. The character in parenthesis after the scene name indicates whether the scene is synthetic (S) or real-world (R).}
}
The results show interesting patterns in the objectionability of different distortions. 
\OPT\ offers a consistent performance improvement over \NN. The only exception is the \emph{Furniture} scene featuring thin and irregularly shaped foreground objects, in which case all types of view interpolation are more objectionable than the selection of the nearest single view. The optical flow interpolation works better for real-world scenes as there are more features that can be detected. The \LINEAR\ interpolation in most of the cases results in the worst performance, except for small distortion levels, which may indicate that visible blur due to this distortion is strongly objectionable. Similar findings have been reported by Vangorp~\etal\cite{Vangorp2011} in their study on the visual performance of view interpolation methods in monocular vision. 
\HEVC\ and \GAUSS\ distortions are usually the easiest to detect as they induce significant amount of spatial distortion when compared to others.
Overall the results show clearly that light-field quality is scene-dependent and successful quality metric must predict the effect of scene content on the visibility of light-field distortions.

\mysection{Evaluation of quality metrics}{Metric}
We considered several popular image, video, stereo, and multiview quality metrics. We briefly describe the metrics and then show their individual performance on our dataset. For obtaining the quality of a light field using image quality metrics, we apply the metrics on individual light-field images and then average the scores over all images.

\textbf{Quality metrics:}
\changed{Although studies show that perceptual metrics perform better than an absolute difference (AD)~\cite{Lin11}, because of its significant usage in image quality assessment, we considered peak signal-to-noise ratio (\PSNR). We also investigated \SSIMSS~\cite{Wang2004}, which is widely used on 2D images, and its extensions to angular domains -- {\SSIMSSA} and {\SSIMSAA}. {\SSIMSAA} computes the same statistics as standard \SSIMSS but on 3D patches extracted from the light-field volume. {\SSIMSSA} uses {2D$\times$1D} patch which contains a 2D window extracted from a particular view and a 1D row of pixels that extends from the center of the 2D window in the angular domain (see Figure{~\ref{fig:2D-3D}}). We applied the metrics to all light-field images without resampling and averaged the scores over all images. Although we experimented with various pooling strategies, we found that the average value performs best.
Due to better performance, we chose the angular window sizes of 32 and 64 pixels for \SSIMSSA\ and \SSIMSAA\ respectively. We also considered a multi-scale version of \SSIMSS -- \MSSIM~\cite{Wang2003} which extends \SSIMSS\ to compute differences on multiple levels. We also used \GMSD~\cite{Xue2014} which provides good performance over a rich collection of image datasets. The most advanced 2D metric considered in our experiments was \VDP~\cite{Mantiuk2011} which stands out among perception-based quality metrics.
}
\myfigure{2D-3D}{Patches used in our extensions of \SSIMSS.}

We further considered the NTIA General Model -- \VQM ~\cite{Pinson2004} which was standardized for video-signals evaluation (ANSI T1.801.03-2003). For this metric, light-field images are input in a form of video panning from the leftmost view to the rightmost view and back. We also chose the stereoscopic image quality metric -- \CHEN \cite{Chen2013} that is based on the concept of cyclopean image where, we averaged scores obtained from all stereo pairs shown in our experiment. To capture the full range of stereo quality metrics, we also included a stereoscopic video quality metric \SILVA~\cite{DeSilva2013}. 

Finally, we chose metrics that address multiview data and account for interpolation artifacts. \BATTISTI\ proposed by Battisti et al.~\cite{Battisti2015} first shift-compensates blocks from the reference and distorted (interpolated) images. These matched blocks undergo the first level of Haar wavelet transform and histogram of the sub-band corresponding to horizontal details in the block is computed. Finally, the  Kolmogorov-Smirnov distance of these histograms is taken as the metric prediction. Another metric for the multiview video is \SANDIC~\cite{Sandic2016}. It computes the multi-resolution morphological pyramid decomposition on the reference and test images. Detail images of the top levels of these pyramids are then compared through the mean squared error. The resulting per pixel errors maps are then pooled and converted to a peak signal-to-noise ratio measure.

\mysubsection{Metric performance comparison}{Metric-performance}

\mycfigure{barplot}{The goodness-of-fit scores for the metrics expressed as Pearson Correlation Coefficient ($\rho$) and reduced chi-square ($\chi^2_{red}$) after cross-validation. The results for each cross-validation fold are shown. $\chi^2_{red} = 1$ indicates that the goodness of fit between the metric predictions and the subjective data is in perfect agreement with the measured subjective variance and $\rho = 1$ indicates perfect positive linear relation between objective scores and JODs. The error bars represent standard error.}
The quality values predicted by each metric are expected to be related to JOD values, but this relation can be complex and non-linear. To account for this relation, we follow a common practice and fit a logistic function:
\begin{equation}
q(o) = a_1 \left\{\frac{1}{2} - \frac{1}{1 + \exp\left[a_2 \left( o -
a_3 \right)\right]}\right\} + a_4 o + a_5\ \quad
\end{equation}
where $o$ is the output of a metric. 
The parameters $a_{1..5}$ are optimized to minimize a given goodness-of-fit measure. We computed
several such measures, such as Spearman rank-order correlation, or
MSE, which can be found in the supplementary materials. Here we report
the reduced chi-squared statistic ($\chi^2_{red}$) and Pearson
correlation coefficient ($\rho$). $\chi^2_{red}$ is computed as a
weighted average of the squared differences, in which weights are the
inverse of sample variance. This accounts for the fact that larger JOD
values are more uncertain (refer to Figure~\ref{fig:exp_results}), and
therefore, the accuracy of their prediction can be lower. For a fair
comparison, we employed a seven fold cross-validation across different
scenes. We measured the goodness-of-fit on two randomly chosen scenes
in a cross-validation fold and averaged the results over all
folds. The resulting Pearson correlation and $\chi^2_{red}$ values are
shown in Figure~\ref{fig:barplot}. The performance of the metrics on
individual distortions are shown in Figure~\ref{fig:distBarPlots}. \changed{A more elaborate analysis including the evaluation on real-world and synthetic scenes separately is presented in the supplementary materials.}

The results show good performance of 2D image and video quality
metrics. This is unexpected as our dataset was meant to emphasize
visibility of angular artefacts, which are not directly considered by
these metrics. We observed, however, that angular distortions
indirectly translate into the differences in spatial patterns, which
could explain the good performance.
We hypothesize that relatively better performance of \VDP and \GMSD is
achieved by detecting changes in contrast across multiple scales,
which in case of \VDP is additionally backed by perceptual scaling of
distortions and discarding of those that are invisible. A comparable
performance of video (\VQM) and stereoscopic (\SILVA) metrics can be
explained by their emphasis on the relation between neighboring views,
which in some way captures angular aspects of light-field viewing.
Figure~\ref{fig:distBarPlots} shows that some metrics are better at
predicting some distortion types than the others. For example, \VDP
consistently under-predicts quality for \HEVC. Training such metrics
for a particular distortion type could substantially boost their
performance. Unexpectedly, our {\it ad hoc} attempts to extend the
\SSIMSS metric by adding the angular dimension (\SSIMSAA) or right
away considering 3D patches (\SSIMSSA) that should account for angular
changes has led to significantly worse results. Clearly, there is a
room for improvements and a suitable dataset, such as the one provided in
this work, should help to develop a better metric in future.
%


\mycfigure{distBarPlots}{ Left: The prediction accuracy per-distortion reported as reduced chi-squared goodness of fit score. Middle and right: $\chi^2_{red}$--fit for the metrics HDR-VDP and GMSD over all scenes. The prediction accuracy for individual distortions are shown inside the plots and the overall accuracy is indicated on the top of the plots.}

\mysubsection{Sparse light-field reference case}{distorted-reference}
\myfigure{dist_ref_plot}{The goodness-of-fit scores for the subset of
  the dataset when a dense LF is used as a reference (blue), when
  nearest-neighbour at the 2$^{nd}$ distortion level is a reference
  (cyan), or when optical flow is used to up-sample the reference
  LFs. The dots at cyan bars mean that the value is statistically
  different from the dense LF case and the dots on the yellow bars that
  the values are statistically different from the \NN case. The
  significance is computed by bootstrapping and running one-tailed
  test ($p=0.05$). }

In all our tests, we provided a high quality, 101-view light field as a
reference for the quality metrics. In practice, in most applications
only sparsely sampled light field is unavailable. When a sparse
light field is used as a reference, a full-reference metric is given
to compare two distorted light fields without a perfect reference. This is a task that such metrics were not designed for as they are intended to predict JODs relative to the perfect reference image, not JNDs relative to any other image (refer to Figure~\ref{fig:jod_vs_jnd}). This issue is
potentially shared with other quality assessment tasks, for example
when a metric is trained on 4K images, but it is used on much lower
resolution images. However, this problem is exacerbated in case of
light fields, where the reduction of angular resolution is often
substantial.

To test whether the metrics can predict the quality of distorted light fields using sparse light fields as a reference, we measured the performance of the metrics on a subset of our dataset. As a reference, we chose light fields with distortion \NN and severity level two, which correspond to original light fields subsampled to 21 angular views. For the testing light field, we considered all light fields with a higher distortion levels. For a fair comparison, we also ran the metrics on the same subset using full 101-view light fields as a reference. The results of these tests are shown as cyan and blue bars in Figure~\ref{fig:dist_ref_plot}. The significant difference in
goodness-of-fit scores (marked with dots) show that metrics predictions get worse if imperfect (sparse) reference is used. This suggests that the existing metrics must be provided with a high-quality reference light field to predict reliably the quality. 

But if such high quality reference is not available, can it be
approximated? Our subjective data shows that optical-flow
interpolation (\OPT) produces the highest quality results. Therefore,
we used \OPT to produce reference 101-view light fields from sparse
21-view light fields and reran the metrics on the subset. The results indicate that
the predictions improved as compared to using sparse light field (yellow
vs. cyan bars in Figure~\ref{fig:dist_ref_plot}). This suggests that a potential solution to the problem of imperfect reference is to use high-quality interpolation method in order to generate reference.

\mysection{Conclusions and future work}{conclusions}
We have established a new 3D dense light-field dataset together with the subjective quality scaling for various distortions that occur in light-field applications. 
Different methods in light-field processing lead to visual artifacts with quite different appearance, e.g., blur for \LINEAR, ghosting for \OPT, image flickering and jumping for \NN. 
Our experiments reveal how these different artifacts affect perceived quality. Our subjective scores are derived from an interactive 3D light-field viewing setup and correspond precisely to overall quality of light fields rather than individual views. We have evaluated the potential of existing image, video, stereo, and multiview quality metrics in predicting the subjective scores. Our observations show that the metrics -- \VDP, \GMSD, \SILVA and \VQM perform reasonably well when comparing a distorted light field to a dense reference, and can be used in applications requiring such comparisons. When dense light field is not available, which is the case in some applications, the usage of these metrics for quality assessment is not justified. The perceptually scaled data that we provide can be used for training and validating new light-field quality metrics. Of practical interest for such development is the problem identified in this work, where incomplete, sparse light fields must serve as the reference. Our results also reveal the quality of different light-field reconstruction method, which can directly guide the choice of the light-field reconstruction technique. In the current work, we did not consider aspects such as masking properties of the human visual system. It could be interesting to investigate how much the metrics gain by considering this effect rather than simple averaging of scores over all views. When creating our dataset, we did not consider focus cue. We are, however, not aware of any display setup that could be used to evaluate both motion parallax and focus cue quality. We also believe that the problems revealed in this work should be addressed before including additional cues. 

\vspace{2ex}
\textit{\textbf{Acknowledgements:}
This project was supported by the Fraunhofer and Max Planck cooperation
program within the German pact for research and innovation (PFI).
Denis Sumin was supported by the European Union's Horizon 2020 research and innovation programme under the Marie Sklodowska-Curie grant agreement No 642841. The authors would like to thank Tobias Ritschel for the initial discussions and providing synthetic scenes.}

{\small
\bibliographystyle{ieee}
\bibliography{Paper}
}
\end{document}